%% file: conference_101719.tex
\documentclass[a4paper, 10 pt, conference]{ieeeconf}
\IEEEoverridecommandlockouts
\usepackage{cite}
\usepackage{amsmath,amssymb,amsfonts}
\usepackage{algorithmic}
\usepackage{graphicx}
\usepackage{textcomp}
\usepackage{xcolor}
\usepackage{comment}
\usepackage{bm}
\def\BibTeX{{\rm B\kern-.05em{\sc i\kern-.025em b}\kern-.08em
    T\kern-.1667em\lower.7ex\hbox{E}\kern-.125emX}}

\input{macros.tex}  
\usepackage{flushend} 
\usepackage{siunitx} 
\usepackage{gensymb} 
\usepackage{multirow} 
\usepackage[tableposition=top, font=small]{caption}
\usepackage{subfig}
\usepackage{colortbl} 
\usepackage{xcolor} 

\usepackage{hyperref}
\hypersetup{%
  colorlinks=true,%
  allcolors=.}

\newcommand{\subfig}[2]{Fig.\,\ref{#1}\,\subref{#2}}

\makeatletter
\def\namedlabel#1#2{\begingroup
    #2%
    \def\@currentlabel{#2}%
    \phantomsection\label{#1}\endgroup
}
\makeatother
    
\begin{document}
\title{\LARGE \bf
Assembling Solar Panels by Dual Robot Arms 
\\
Towards Full Autonomous Lunar Base Construction
}

\author{Luca Nunziante$^{*1,2}$, Kentaro Uno$^{2}$, Gustavo H. Diaz$^{2}$,\\
Shreya Santra$^{2}$, Alessandro De Luca$^{1}$ and Kazuya Yoshida$^{2}$ 
\thanks{$^{1}$L. Nunziante and A. De Luca are with the Department of Computer, Control and Management Engineering, Sapienza University of Rome, Italy. This work was performed while L.\ Nunziante was visiting the Tohoku University.
 Email: nunziante.2015361@studenti.uniroma1.it, deluca@diag.uniroma1.it}%
\thanks{$^{2}$L. Nunziante, K. Uno, G.H. Diaz, S. Santra, and K. Yoshida are with the Space Robotics Lab. (SRL) in Department of Aerospace Engineering, Graduate School of Engineering, Tohoku University, Sendai 980-8579, Japan.}%
\thanks{\textit{$^{*}$Corresponding author is Luca Nunziante.}}
}%

\maketitle

\begin{abstract}
Since the successful Apollo program, humanity is once again aiming to return to the Moon for scientific discovery, resource mining, and inhabitation.
Upcoming decades focus on building a lunar outpost, with robotic systems playing a crucial role to safely and efficiently establish essential infrastructure such as solar power generating towers.
Similar to the construction of the International Space Station (ISS), shipping necessary components via modules and assembling them in situ should be a practical scenario.
In this context, this paper focuses on the integration of vision, control, and hardware systems within an autonomous sequence for a dual-arm robot system. We explore a perception and control pipeline specifically designed for assembling solar panel modules, one of the benchmark tasks. Ad hoc hardware was designed and tested in real-world experiments. A mock-up of modular solar panels and active-passive connectors are employed, with the control of this grappling fixture integrated into the proposed pipeline. The successful implementation of our method demonstrates that the two robot manipulators can effectively connect arbitrarily placed panels, highlighting the seamless integration of vision, control, and hardware systems in complex space applications.
\end{abstract}




\section{Introduction}
In the past years, the interest in lunar exploration has been growing worldwide~\cite{Creech2022_Artemis}. Up-to-date successful missions include the Indian Space Research Organization (ISRO)'s Chandrayaan-3, Japan Aerospace Exploration Agency (JAXA)'s extremely precise soft landing on the moon with the SLIM (smart lander for investigating moon) mission, and a USA based startup, Intuitive Machines's IM-1 lunar landing~\cite{Nikkei2024}. From navigating the challenging lunar terrain to conducting complex manipulation tasks, autonomous robotic systems offer a unique set of capabilities that have the potential not only to boost scientific progress but also to mitigate risks associated with human missions. 
To ensure self-sustenance and to support human presence on the Moon, robots capable of autonomous manipulation tasks are particularly relevant. During the early stages of lunar base development, these robots must create vital infrastructure such as solar power units and communication stations with Earth. A practical approach is to ship all required components in modules and have the robots assemble them on-site. In this scenario, the robots should also be modular.
\begin{figure}[t]
\centering    
    \includegraphics[width=\linewidth]{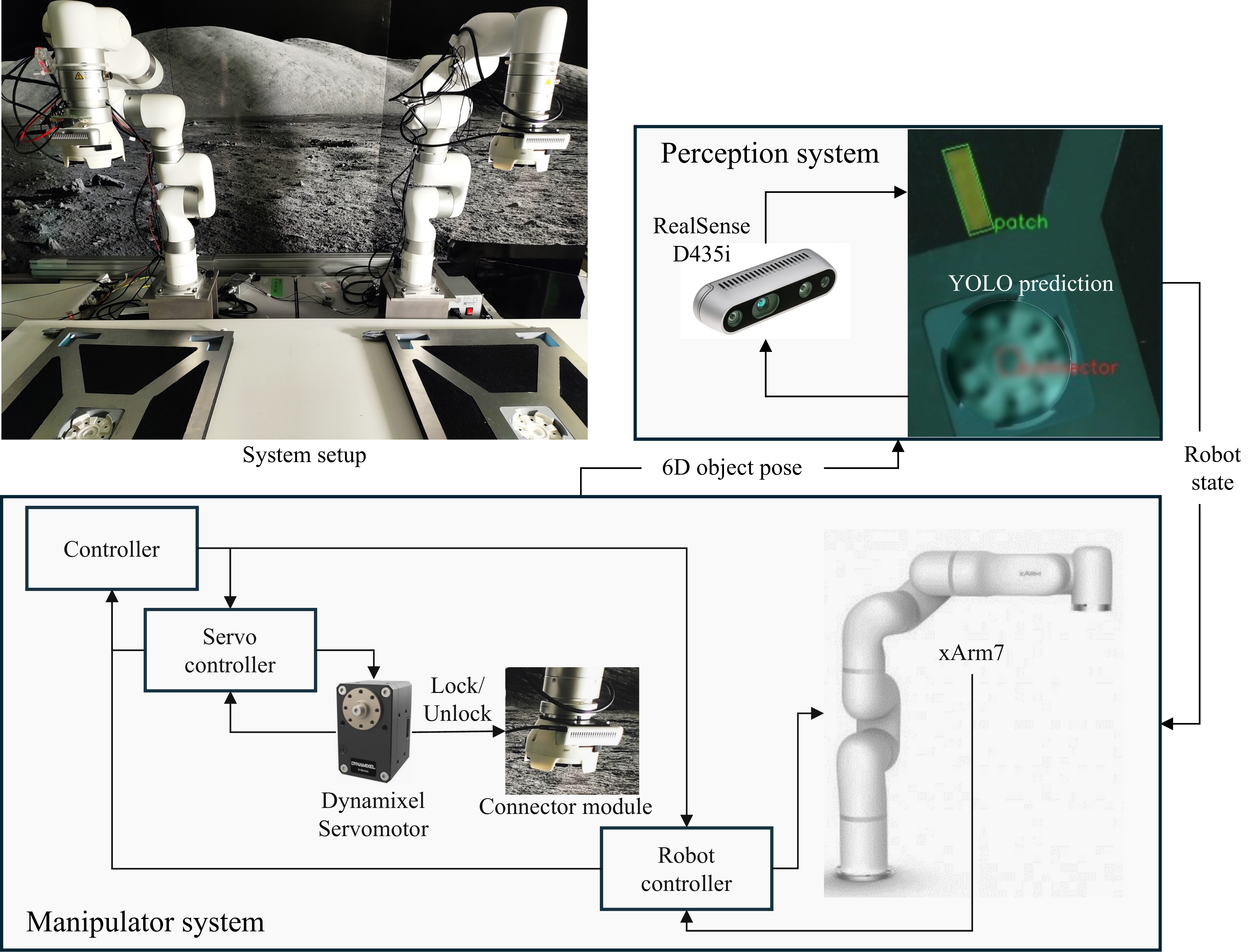}
    \caption{Overall hardware setup (top) and functional scheme to illustrate the integration and interaction of the perception and control modules (bottom).}
    \label{fig:figure1}
    \vspace{-15pt}
\end{figure}

In this work we present a perception and control pipeline to perform autonomous assembly tasks in a multi-robot setup; two robot arms have the task to localize two solar panels, pick them up with dedicated connectors, safely lift them avoiding collision with the table they are placed on and ultimately assemble them. 
The perception module is a YOLOv8.1~\cite{Jocher2023_Ultralytics_YOLO} (You Only Look Once) model, an extension of YOLOv8 trained to predict Oriented Bounding Boxes. The control module integrates data coming from the vision system with classical control methods like impedance control, Nonlinear Model Predictive Control (NMPC) and force control~\cite{Siciliano2010}, showcasing a comprehensive system integration. The major contributions of this paper are highlighted as follows:
\begin{itemize}
    \item Fully autonomous pipeline of the robotic assembly of the solar panels, which is regarded as the essential milestone task, is designed and implemented.
    \item Hardware and software systems are optimally integrated, resulting in  successful real-world task demonstration.
\end{itemize}
The main goal of this work is to present a fully autonomous pipeline that integrates a state-of-the-art perception module with the control of novel hardware in a multi-robot cooperative scenario.

The remaining part of this paper is structured as follows. In Section \ref{sec:relatedworks} we review the state-of-the-art perception and control schemes as standalone components. In Section \ref{sec:method} we design the aggregation of the modules and the overall pipeline for the panels assembling. Section \ref{sec:experiments} details the experimental setup to conduct the robotic demonstration considering the actual task scenario. Section \ref{sec:conclusions} finally summarizes the results and discusses future work.

\vspace{-3pt}
\section{Related Work}\label{sec:relatedworks}
In this work, object detection, impedance control, force control, and Nonlinear Model Predictive Control (NMPC) are exploited to design a pipeline to achieve full autonomous assembly of structure modules. The relevant background on these techniques is recalled hereafter.

\vspace{-5pt}
\subsection{Perception}
Object detection is a computer vision task aimed at localizing and identifying objects within an image, assigning a label to each identified object. Thanks to the seminal work~\cite{Krizhevsky2012_CommACM2017} that introduced Deep Convolutional Neural Networks (DCNNs) into the computer vision field, and to the technological advancements of the last decades, DCNNs undoubtably became the standard for computer vision tasks at large.

In the evolution of DCNNs for object detection, two branches can be identified: two-stage and one-stage detectors~\cite{Xiao2020}. The models of the former category separate the object location task and the object classification task, while models of the latter generate the class probabilities and location coordinates of an object in a single stage. Separating the two tasks yields accurate models with slower inference speed~\cite{Lin2017_CVPR} which may not be suitable for real-time applications, while one-shot models trade off higher speed for lower accuracy.

Through the years, the one-shot YOLO series stood out as a standard for real-time object detection thanks to the high speed and accuracy achieved. Upon its release, YOLO was the first one-stage model for object detection, boasting impressive inference speed but suffering in terms of localization accuracy~\cite{Redmon2016_YOLO}. Over the subsequent versions much improvement was made, until the latest release which is YOLOv9~\cite{Wang2024_yolov9}. YOLO models are extremely versatile and have been applied to a wide array of different tasks, including but not limited to construction: building classification~\cite{Khatua2024}, healthcare: fracture detection~\cite{Chien2024_fracture} and agriculture: fruit ripeness identification~\cite{Xiao2024}.


\subsection{Control}
Impedance control is a control method that makes the robotic system react in a desired way to external force, imposing a desired dynamic behavior to the interaction between the robot's end-effector and the environment. The impedance model imposed is
\begin{equation}
    \bm{M}_{\mathrm{m}}(\ddot{\bm{r}}-\ddot{\bm{r}}_\mathrm{d})+\bm{D}_{\mathrm{m}}(\dot{\bm{r}}-\dot{\bm{r}}_\mathrm{d})+\bm{K}_{\mathrm{m}}(\bm{r}-\bm{r}_\mathrm{d})=\bm{f}_{\mathrm{e}}
    \label{eq:impedance_model}
\end{equation}
where $\bm{r} \in \mathbb{R}^r$ is a representation for the relevant components of the end-effector pose (position and orientation), $r$ is the dimension of the robot task --- e.g., three for planar tasks, six for spatial --- and $\bm{r}_{\mathrm{d}}$ is a desired pose. ${\bm{M}_{\mathrm{m}} >0 ,\bm{D}_{\mathrm{m}} \geq 0 ,\bm{K}_{\mathrm{m}} > 0 \in \mathbb{R}^{r \times r}}$ are respectively the desired mass, damping and stiffness imposed by control, and ${\bm{f}_{\mathrm{e}} \in \mathbb{R}^{r}}$ are the generalized external forces applied on the end-effector. With this scheme, it is possible to indirectly control the contact forces, e.g., to prevent damaging impacts in case of environment uncertainties or to assign a soft environment interaction. Thanks to such versatility, this control scheme is widely used not only in mechanical manipulation but also in human-machine interaction and motion of robotic devices with adjustable compliance~\cite{Song2019}.

Unlike impedance control, force control schemes are aimed at precisely regulating a desired contact force at the end-effector level where contact is expected. Thus, differently from impedance control where a desired end-effector pose $\bm{r}_{\mathrm{d}}$ is used, in this scheme a desired force reference is required. To reduce the dynamics of the manipulator to a free-floating mass, a Cartesian feedback linearization law can be employed first~\cite{Isidori1995}. It is then possible to use linear, lumped parameter models of different order to analyze the interaction between the robot arm, the force sensor at its end-effector, and the environment~\cite{Eppinger1986}. In this way, simple linear controllers (e.g., PID or one of its subsets) can be implemented to regulate efficiently the contact force. 

Model Predictive Control (MPC) is a control strategy that operates by repeatedly solving a constrained Optimal Control Problem (OCP) over a finite time horizon, using a model of the plant to predict future behavior. At each control sampling instant an optimal control sequence that maximizes a performance measure, which usually takes the form of a cost to be minimized, is found and the first element of the control sequence is issued. At subsequent control instants, the process is repeated using the latest available information on the system state. This introduces feedback into the control scheme and robustness to uncertainties or unknown variations in the model~\cite{Kouvaritakis2016}. In general, one distinguishes between linear and nonlinear model predictive control (NMPC). In the former case, linear models are used for the system dynamics, states and inputs are subject to linear constraints, and a quadratic cost function is used; instead, NMPC refers to MPC schemes that are based on nonlinear models, and/or consider non-quadratic cost functionals and/or impose general nonlinear constraints on the states and inputs~\cite{Findeisen2003}. Due to its robustness and ability to handle constraints and general nonlinear systems, MPC has been widely used in process control in industries both in its linear~\cite{Qin1997} and nonlinear formulation~\cite{Findeisen2002}.

\vspace{-5pt}
\section{Method}\label{sec:method}
The proposed method integrates the previously outlined perception and control methodologies into an assembly pipeline. \fig{fig:flowchart} illustrates a flowchart delineating the various phases of the assembly process. In this section, we elaborate on each element within this scheme, whose structure is reminiscent of the Finite State Machine developed in \cite{Diaz2024_AROB}.

\vspace{-5pt}
\subsection{Approaching and Grasping}
\begin{figure}[t]
\centering    
    \subfloat{
    \includegraphics[width=0.47\linewidth]{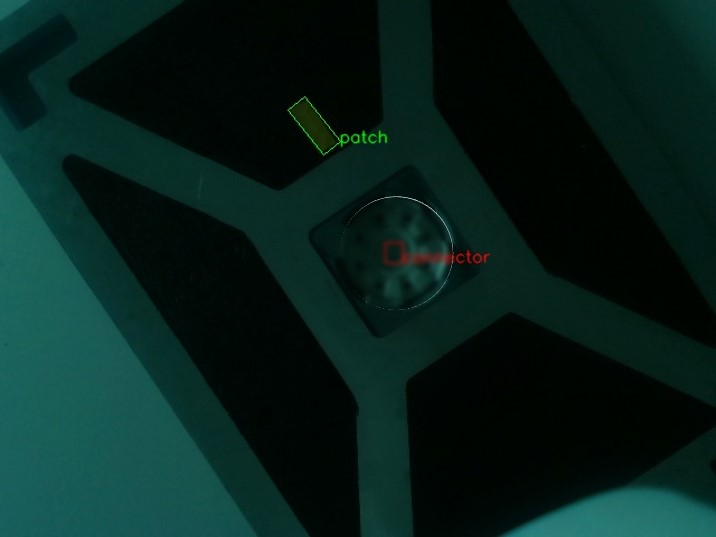}
    }
    \subfloat{
    \includegraphics[width=0.47\linewidth]{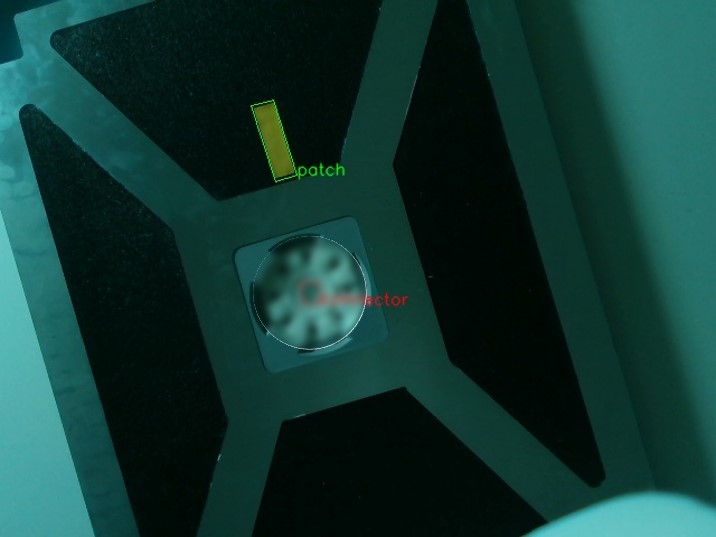}
    }
    \caption{Result of the inference made by the YOLO model in challenging lighting conditions: dark assumed in Lunar environment. From the patch detection we retrieve the panel orientation, while the connector detection provides the desired position for the end-effector.}
    \label{fig:yolo_obb}
    \vspace{-20pt}
\end{figure}

The perception module comprises a YOLOv8.1 model and a deprojection operation. As the YOLO model performs inference only once in the initial state, high inference speed is not a requirement and the largest model with 69.5M parameters can be used. The network is pretrained on the DOTA-v1.0~\cite{Xia2018} dataset and fine-tuned on a custom dataset made of 309 training images and two classes. Inference results in challenging lighting conditions are shown in \fig{fig:yolo_obb}.

From the detection of the patch we retrieve the orientation of the panel on the table, and from the pixel coordinates of the connector, using the depth information from the stereo camera mounted on the end-effector, we deproject the point first from the pixel space to the 3D camera frame, and finally to the robot base frame using the robot state. This procedure is analogous to the one in~\cite{Boucher2024}. The deprojected point together with the panel orientation retrieved from the patch detection constitute the desired pose $\bm{r}_\mathrm{d} \in \mathbb{R}^6$ that is used in \eqref{eq:impedance_model}, where we can further specify the approach velocity and acceleration $\dot{\bm{r}}_\mathrm{d},\ddot{\bm{r}}_\mathrm{d}$. This impedance-based approach will allow the correct grasp of the panel even in the presence of small detection errors that may arise, especially in challenging lighting conditions typical of lunar scenarios (see \fig{fig:yolo_obb}).

Once the active side of the grappling fixture is properly in place (see \fig{fig:hardwareSetup}), we command the servo motor to lock the mechanism, thereby attaching it to the passive side and securing the grasp.

\subsection{Collision-free lifting}
\begin{figure}[t]
\centering
    \includegraphics[width=\linewidth]{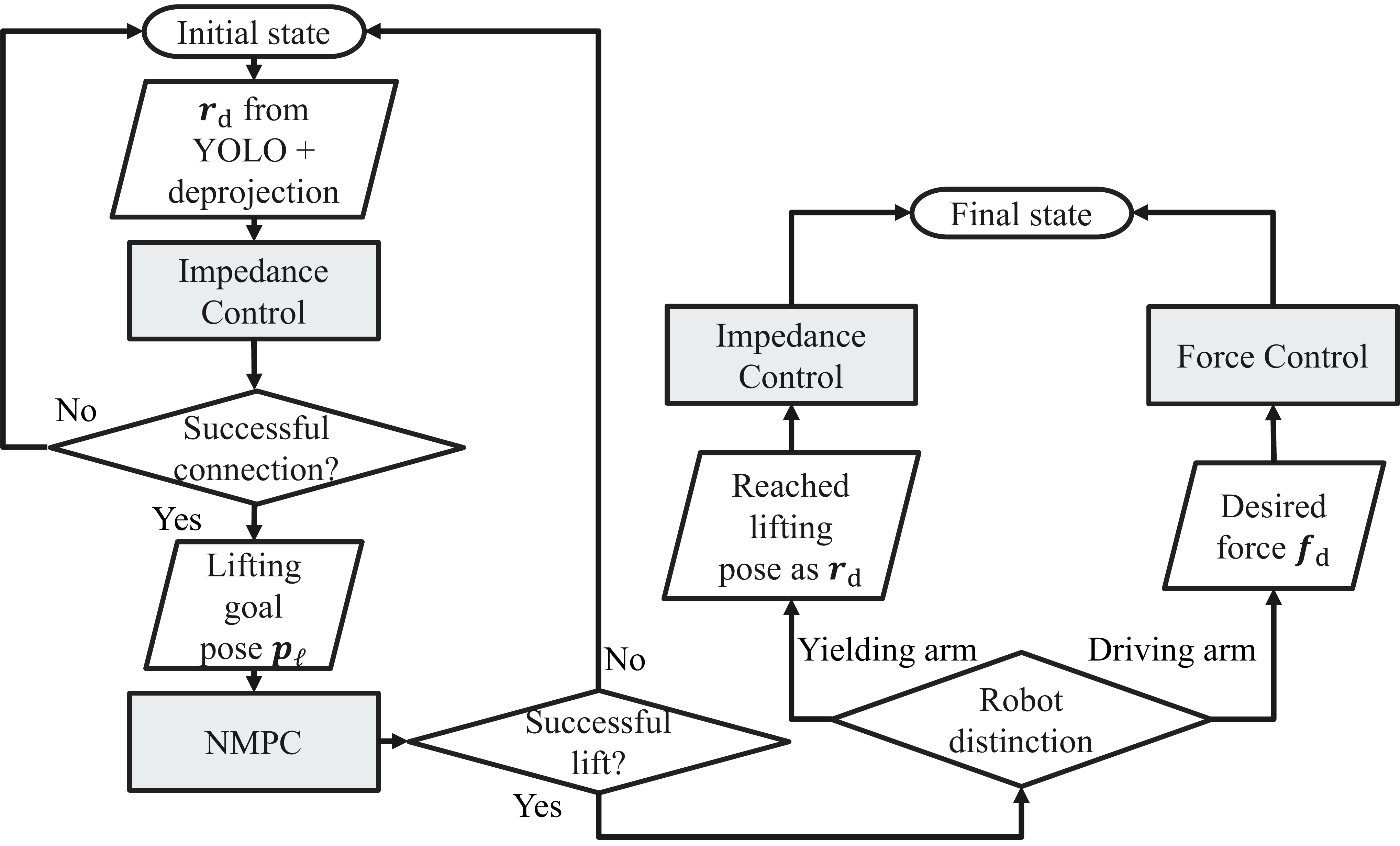}
    \caption{Flowchart of the assembly pipeline. Impedance control is utilized to adjust to the pushed force when grasping the panels as well as inserting one panel into the other. NMPC allows controlling the panel picking up trajectory avoiding the collision with the surrounding environment.}
    \label{fig:flowchart}
    \vspace{-18pt}
\end{figure}

After the successful connection, the robot enters in velocity control mode to bring the end-effector to a desired pose where the panels are lifted. Since the payload is now attached to the end-effector, it is crucial to avoid collisions between the payload, i.e., a panel and obstacles in the environment, such as the table on which it is placed. The choice of NMPC, differently from other control choices, allows to explicitly consider nonlinear collision avoidance constraints, offering also the additional flexibility of dynamically changing constraints in non-static environments. 

In this framework, we consider the end-effector position to be directly controllable via velocity commands, while for the orientation we adopt an ingenious approach. Given the current orientation represented by the rotation matrix ${\bm{R}_\mathrm{A} \in SO(3)}$ and the desired one represented by ${\bm{R}_\mathrm{B}\in SO(3)}$, where $SO(3)$ is the special orthogonal group of order 3, we determine a unit axis ${\bm{a} \in \mathbb{R}^3}$ and an angle ${\theta_{\mathrm{AB}} \in \mathbb{R}}$ such that
\begin{equation}
    \bm{R}(\bm{a},\theta_\mathrm{AB})=\bm{R}_\mathrm{A}^\top\bm{R}_\mathrm{B} \ .
    \label{eq:relativerotation}
    \vspace{-5pt}
\end{equation}
Using the axis-angle representation with $(\bm{a},\theta_\mathrm{AB})$ has two main advantages:
\begin{itemize}
    \item Constraining the change of orientation from $\bm{R}_\mathrm{A}$ to $\bm{R}_\mathrm{B}$ to occur as a rotation around a predefined axis offers more predictable intermediate orientations than considering the elements of an Euler representation independently. This is not a secondary aspect when dealing with large payloads, as in our application.
    \item Within the NMPC scheme, a single scalar variable $\theta$ is used to represent the orientation. The value of $\theta$ will be 0 at the initial instant and has to reach to the desired value of $\theta_\mathrm{AB}$.
\end{itemize}
Thus, the desired end-effector pose ${\bm{r}_\mathrm{d}\in\mathbb{R}^6}$ is transformed to a lifting goal ${\bm{p}_\ell=(x_d,y_d,z_d,\theta_\mathrm{AB}) \in \mathbb{R}^4}$ where $(x_d,y_d,z_d)$ is the desired end-effector position and $\theta_{\mathrm{AB}}$ comes from \eqref{eq:relativerotation}.

Initially, we assume to be able to control the value of $\theta$ directly via a velocity input $u_\theta$. Therefore, the state of the system used in the NMPC scheme is ${\bm{x}=\begin{pmatrix}x & y & z & \theta\end{pmatrix}\in\mathbb{R}^4}$, the state space representation is
\begin{equation}
        \dot{x}=u_{x}, \quad \dot{y}=u_{y}, \quad \dot{z}=u_{z}, \quad \dot{\theta}=u_\theta
    \label{eq:linear_system}
\end{equation}
and we formulate the OCP to solve at each control instant as
\setcounter{equation}{4} 
\begin{IEEEeqnarray}{lll}
\IEEEyessubnumber*
    \min_{\bm{x}(\cdot),\bm{u}(\cdot)} \int_0^T\|\bm{x}(\tau)-\bm{p}_\ell\|^2_{\bm{Q}} +\|\bm{u}(\tau)\|^2_{\bm{R}}\mathrm{d}\tau \nonumber\\
    \quad \quad \quad + \> \|\bm{x}(T)-\bm{p}_\ell\|^2_{\bm{W}^e} \label{eq:costfunc}\\
    \text{subject to:} \nonumber \\
    \bm{x}(0)=\bm{x}_\mathrm{0} \label{eq:initialstate}\\
    \bm{f}(\bm{x}(t),\dot{\bm{x}}(t),\bm{u}(t))=\bm{0} & t \in [0,T) \label{eq:dynamics} \\
    \underbar{$\bm{h}$} \leq \bm{h}(\bm{x}(t),\bm{s}) \leq \overline{\bm{h}} & t \in [0,T] \label{eq:feasiblestate}\\
    \underbar{$\bm{u}$} \leq \bm{u}(t) \leq \overline{\bm{u}} & t \in [0,T) \label{eq:feasibleinput}
\end{IEEEeqnarray}
The cost function has a quadratic cost to go and terminal cost to bring the state to the desired value, while also minimizing the control effort. Moreover, $T \in \mathbb{R}^+$ is the prediction horizon, ${\bm{Q} \geq 0},{\bm{R} > 0},{\bm{W}^e \geq 0}$ are weighting matrices of appropriate size,  ${\bm{f} : \mathbb{R}^4 \times \mathbb{R}^4 \times \mathbb{R}^4 \mapsto \mathbb{R}^4}$ is the implicit linear system dynamics from \eqref{eq:linear_system}, $\bm{x}_\mathrm{0}$ is the initial system state, $\bm{s} \in \mathbb{R}^{n_\mathrm{s}}$ is a vector of system parameters, ${\bm{h} : \mathbb{R}^4 \times \mathbb{R}^{n_\mathrm{s}} \mapsto \mathbb{R}^{n_\mathrm{c}}}$ is the nonlinear constraint function imposing a number of $n_\mathrm{c}$ constraints for collision avoidance, $\underbar{$\bm{h}$},\overline{\bm{h}} \in \mathbb{R}^{n_\mathrm{c}}$ are lower and upper bounds for the vector function $\bm{h}$, while $\underbar{$\bm{u}$},\overline{\bm{u}} \in \mathbb{R}^4$ are input bounds. The only source of nonlinearity in this NMPC scheme comes from the collision avoidance constraints imposed in \eqref{eq:feasiblestate}.

Functional for collision avoidance is the set of parameters $\bm{s}$ that includes a representation of the initial orientation of the end-effector $\bm{R}_\mathrm{A}$, the neutral axis $\bm{a}$ from \eqref{eq:relativerotation}, and information about the payload geometry. 
To ensure a collision-free motion, we use the four corners of the payload as control points. Using the information stored in $\bm{s}$, together with the current state of the system, it is possible to retrieve their coordinates in the robot base frame (see \fig{fig:hardwareSetup}). Observing that at any time instant $t$ the actual end-effector (EE) orientation is given by 
\begin{equation}
    \bm{R}_\mathrm{EE} (t)= \bm{R}_\mathrm{A}\bm{R} \left( \bm{a},\theta(t) \right)
    \label{eq:ee_orientation}
\end{equation}
and using knowledge about the payload geometry, it is finally possible to expand the constraints \eqref{eq:feasiblestate} as
\begin{equation}
\begin{aligned}
    v_{x,i} \geq {b}_{\mathrm{collision}}, \quad v_{y,i} \geq {y}_{\mathrm{wall}}, \quad v_{z,i} \geq {z}_{\mathrm{min}},
\end{aligned}
\label{eq:constraints}
\end{equation}
for $i=1,\dots,4$, 
where $\bm{v}_{i}=\begin{pmatrix} v_{x,i} & v_{y,i} & v_{z,i} \end{pmatrix}$ is the $i$-th control point expressed in the robot base frame. The numbering convention begins with $\bm{v}_\mathrm{1}$ representing the top right corner of the panel from a top view (see bottom right image of \fig{fig:hardwareSetup}), and subsequent numbering progresses counterclockwise. Moreover, ${b}_{\mathrm{collision}}>0$ is a constant to avoid collisions of the panel with the base of the robot, ${y}_{\mathrm{wall}}$ erects a virtual wall separating the two robot arms to prevent collision between the payloads during the lift, and ${z}_{\mathrm{min}}$ is the minimum height for not colliding with the table. Differently from the previous two, this last value is not predetermined, but obtained  by recording the end-effector $z$-coordinate when the panel is grasped. Finally, note that the nonlinearity of the constraints \eqref{eq:constraints} with respect to the NMPC state $\bm{x}$ lies in the transformation needed to obtain $\bm{v}_{i}$ in the robot arm base frame $\Sigma_B$.

\subsection{Assembly}
After the robot end-effector reaches the desired pose avoiding collisions, the two arms need to collaborate to successfully assemble the solar panels. To achieve this they are controlled differently: one arm (Yielding arm in \fig{fig:flowchart}) enters again into impedance control mode having a desired pose $\bm{r}_\mathrm{d}$ that is the one reached at the end of the lifting phase, while the other arm (Driving arm in \fig{fig:flowchart}) enters force control mode applying a desired wrench (force and moment) $\bm{f}_\mathrm{d} \in \mathbb{R}^6$ at the end-effector level. Thus, the first arm will be stationary and compliant to external forces according to \eqref{eq:impedance_model}, with these external forces being explicitly applied by the Driving arm. This force interaction leads to a successful assembly, compensating for small uncertainties and pose errors.

\section{Real World Demonstration}\label{sec:experiments}
\begin{figure}[t]
\centering
    \includegraphics[width=\linewidth]{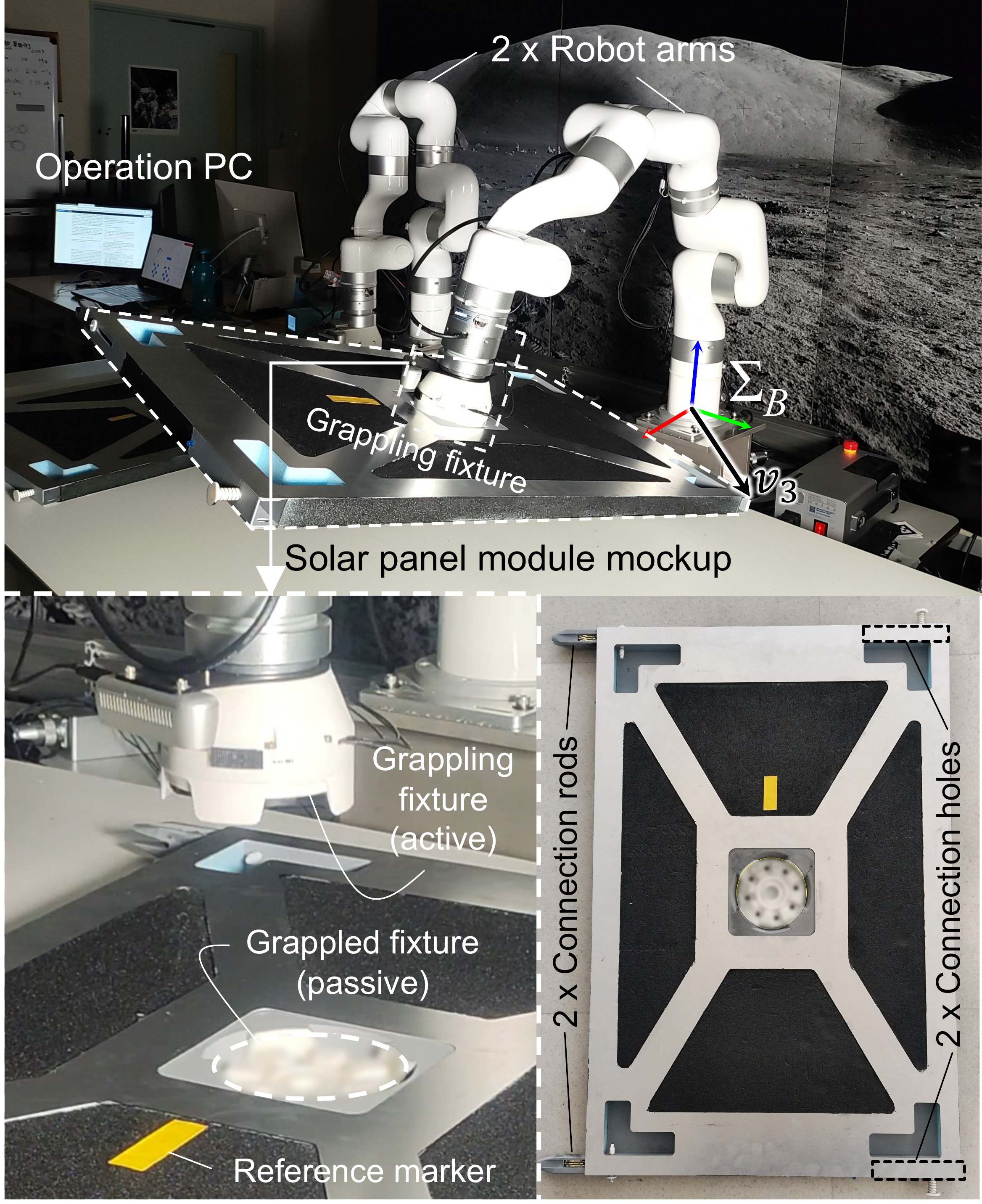}
    \caption{Experimental setup for solar panels assembly demonstration (top), showing the representation of the robot base reference frame and of the control point $\bm{v}_{\mathrm{3}}$: left bottom corner of the panel. The end-effector of the robot has the dedicated grappling fixture to grasp the adapter located at the center of the backside of the panel (bottom left). Each panel has two connection rods at two corners which are inserted into the holes prepared on the other edge for assembly (bottom right).}
    \label{fig:hardwareSetup}
    \vspace{-12pt}
\end{figure}
\begin{figure*}[t]
  \centering
\includegraphics[width=\linewidth]{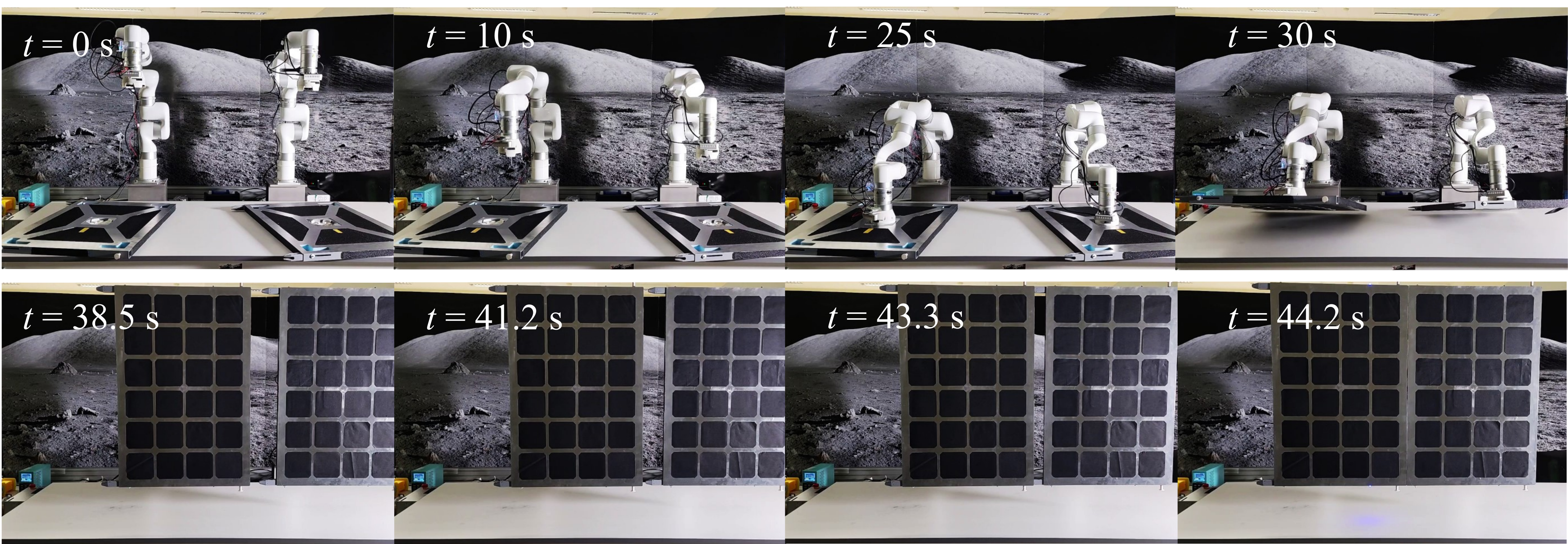}
  \caption{Snapshots of the two panels assembling demonstration by dual robot manipulators in real world. End-effectors are approaching with visual feedback control the detected grappled fixture ($t = 10$\;s), grasping them ($t = 25$\;s), picking the panels up avoiding collision with NMPC ($t = 30$\;s). Then, after achieving the lift-up positions ($t = 38.5$\;s), the right panel is pushed into the other one with force control ($t = 41.2$\;s), and thanks to impedance control the panel is compliant to the applied force ($t = 43.3$\;s), eventually succeeding in the two panels connected ($t = 44.2$\;s). The implemented autonomous sequence was successful irrespectively of the panels' initial pose.}\label{fig:experimentalResult}
  \vspace{-10pt}
\end{figure*}
\begin{figure*}[t]
\centering    
    \subfloat[Position profile of the Yielding arm's end-effector expressed in the base frame $\Sigma_B$]{
\includegraphics[width=0.48\linewidth]{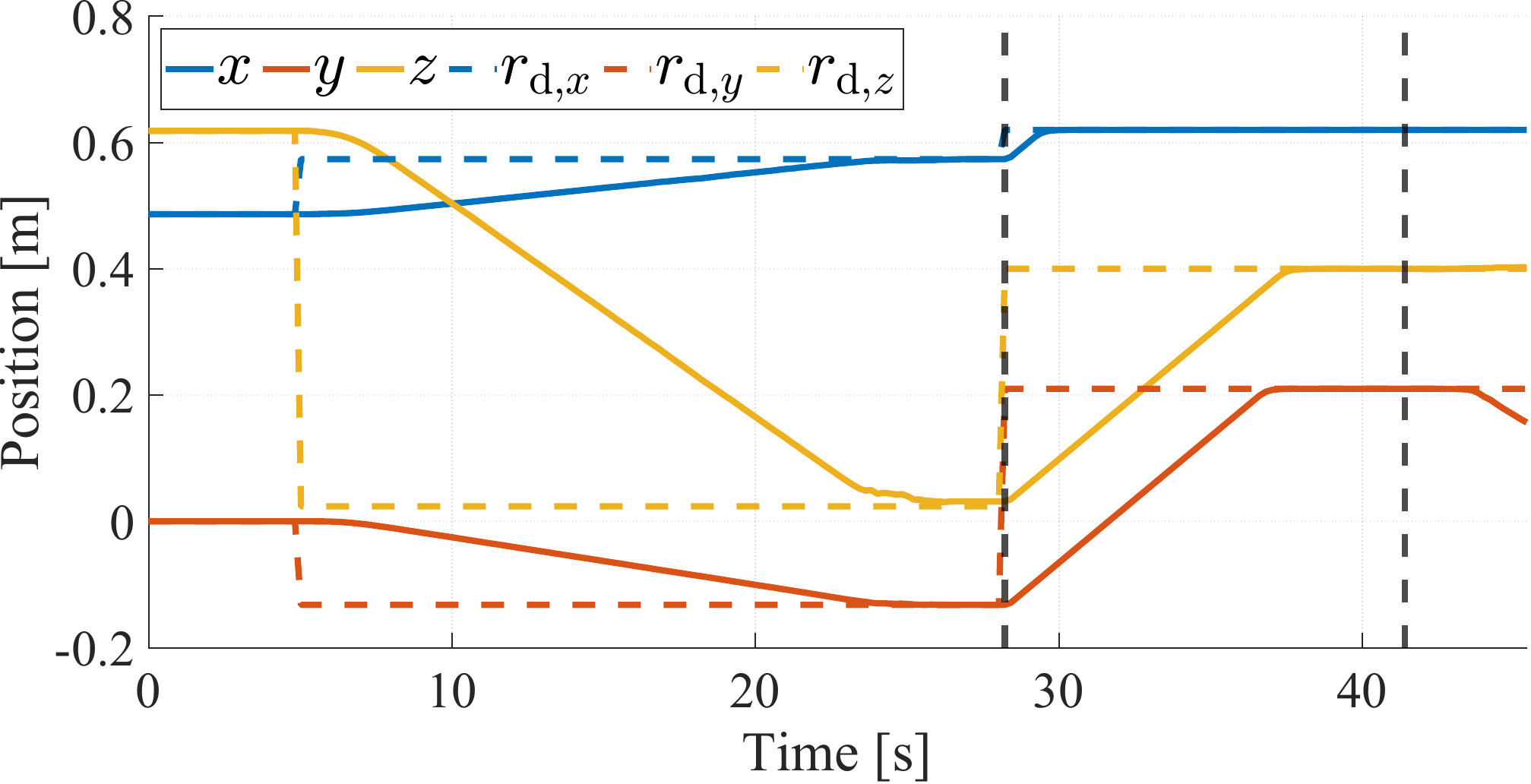}
    \label{fig:plotPos}
    }
    \hfill
    \subfloat[Force measured by the F/T sensor at the tip of the Driving arm expressed in the sensor frame $\Sigma_S$.]{
    \includegraphics[width=0.48\linewidth]{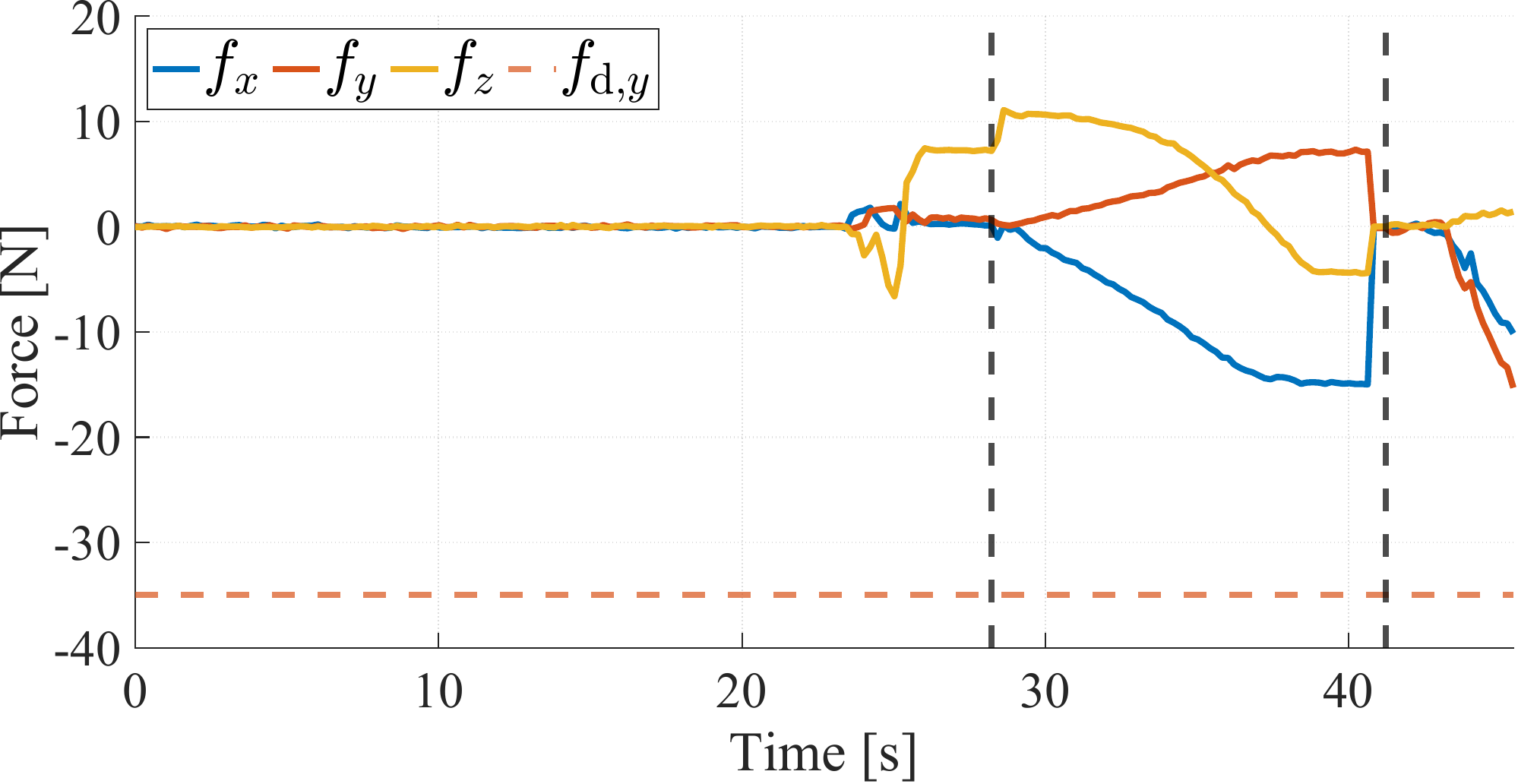}
    \label{fig:plotForce}
    }
    \caption{The vertical black dashed line represents the phase distinction: at $t = 28.2$\;s the lifting phase starts, while at  $t = 41.2$\;s the driving arm starts applying a desired $\bm{f}_{\mathrm{ref}}$ at the end-effector level. At $t = 41.2$\;s we reset the zero of the sensor so that we get rid of the effect of the panel. Contact between the two panels happens at around $t = 43.3$\;s. }
    \label{fig:plotsExperiment}
    \vspace{-10pt}
\end{figure*}
\subsection{Hardware Setup}\label{sec:setup}
The setup consists of the following elements (also see \fig{fig:hardwareSetup}). Two solar panels and two fixed 7-DoF robotic arm manipulators, xArm7 from UFactory, whose end-effector is equipped with an Intel Realsense D435i RGBD camera, a 6-axis force/torque (F/T) sensor, and the active side of the novel grappling modules we present.

The grappling module is made up of an active fixture that is mounted after the F/T sensor on each arm, paired with a passive frame embedded in the center of the panel's backside. The active fixture features a servo actuator that locks the connection upon contact with the passive fixture. Both modules are designed with trapezoidal-shaped guides around the outer frame to help compensate for misalignments between the adapters to a certain extent.

Finally, each panel has two connection rods at the upper corners, which are inserted in the holes designed on the other side. The panels are kept lightweight to be withing the payload capacity of the robot arms, weighing \SI{1}{\kg} each.

\subsection{Results and Analysis}\label{sec:results}
Until now, the input to the system was assumed as the one from \eqref{eq:linear_system}. However, while the desired linear end-effector velocity can be directly issued to the xArm7, $u_\theta$ is not an available input. Since the available input to affect the end-effector orientation is the rate of change of the euler angles, we apply the following mapping
\begin{equation}
    \bm{u}_\Phi=\bm{R}_{\mathrm{A}}\bm{a}u_\theta
    \label{eq:inputMapping}
\end{equation}
and the new input $\bm{u}=\begin{pmatrix}u_{x} & u_{y} & u_{z} & \bm{u}_\Phi\end{pmatrix} \in \mathbb{R}^6$ is issued to the robot.

A representative demonstration of the task of two panels assembly is showcased in \fig{fig:experimentalResult}. Analyzing the force profile measured by the F/T sensor mounted on the Driving arm reported in \subfig{fig:plotsExperiment}{fig:plotForce}, it is possible to recognize the contact between the arm and the panel at $t = 24$\; s due to forces arising particularly along the $z$ axis of $\Sigma_S$. These forces depend on how far off the connector is from the right position, but adjusting the parameters $\bm{M}_\mathrm{m},\bm{D}_\mathrm{m},\bm{K}_\mathrm{m}$ in \eqref{eq:impedance_model} also affects their magnitude. At $t = 41.2$\;s the Driving arm starts applying the wrench $\bm{f}_\mathrm{d}$ at the end-effector that, expressed in $\Sigma_B$, is $\bm{f}_{\mathrm{d}}=\begin{pmatrix} * & -35 & 0 & * & * & *\end{pmatrix}$ where $*$ means that the robot will act rigidly in response to stimuli in those directions. Simultaneously, the Yielding arm will be compliant to external wrenches except along the $x$ axis and around the $z$ axis of $\Sigma_B$, which is equivalent to choosing the corresponding entries in $\bm{K}_\mathrm{m}$ as $+\infty$.  The two panels come in contact when the $y$ coordinate of the Yielding arm's end effector starts changing in \subfig{fig:plotsExperiment}{fig:plotPos} at $t = 43.3$\;s, and the successful assembly is reached when the connecting rods slip into the connecting holes at $t = 44.2$\;s, recognizable in \subfig{fig:plotsExperiment}{fig:plotForce} as a temporary decrease in the measured $f_y$.

To measure the effectiveness of the proposed method quantitatively, in Table \ref{tbl:performance} we report the performance over 40 trials. The main factor that affects these results is a noisy depth estimation, leading to a failed grasp. Improving just this aspect, e.g., by improving the depth estimation from the Intel camera~\cite{rijal2023comparing} or by using a more accurate 6D perception system~\cite{jin2022dope++}, will bring a major increase in the pipeline success rate.
\begin{table}[t]
\centering
\caption{Performance of the proposed pipeline over 40 trials.}
\begin{tabular}{cc|cc}
\hline
Success rate         & Failure rate         & Failure modality                                                & \begin{tabular}[c]{@{}c@{}}Failure rate\\ per modality\end{tabular} \\ \hline
\multirow{2}{*}{0.61} & \multirow{2}{*}{0.39} & Failed grasp                                                    & 0.66                                                                 \\
                     &                      & \begin{tabular}[c]{@{}c@{}}Failed panels\\ insertion\end{tabular} & 0.34                                                                 \\ \hline
\end{tabular}
\label{tbl:performance}
\vspace{-15pt}
\end{table}

\vspace{-2pt}
\subsection{Discussions}\label{sec:discussions}
Together with depth estimation, the other key element for the performance of this pipeline concerns different choices of compliance and rigidity. Beside those discussed in Sec.~\ref{sec:results}, also other combinations have been tested: performance improved while allowing compliance along more axes. Furthermore, during the final assembly phase, we experimented with alternative controller pairs, with some pairs being more favourable. A more sistematic approach to find the right stiffness and damping values in the impedance controller, as well as the force value in the force controller, might improve the success of the overall pipeline.

Having both arms in impedance control required to craft a final goal position for the Yielding arm which would lead to contact between the panels: this is not in the nature of the task as the final assembly position varies depending on the exchanged forces.
Having both arms in force control, the ability to control the nature of the interaction between the panels is lost, differently from what it is possible with impedance control, although the task of autonomous assembly can be done with comparable efficacy. 
Finally, having the Yielding arm in position control compromises the safety of the entire system, as small uncertainties and errors during insertion are met with complete rigidity on one side.
\vspace{-2pt}
\section{Conclusions}\label{sec:conclusions}
In this work, we developed a fully autonomous pipeline for solar panel assembly. The proposed method employs the YOLO visual perception algorithm together with depth information to extract the six dimensional pose of the detected object; then, a nonlinear Model Predictive Control scheme with a minimalist state representation is used for collision avoidance during motion; finally, to accommodate for uncertainties and errors, a combination of impedance and force control makes up the insertion phase. This holistic integration of vision, control, and specialised hardware demonstrates a robust and effective approach to complex assembly tasks in a multi-robot system.

The future scope of this work includes a learning-based approach to evolve between different states and control modes, and implement a method to address the noisy depth estimation limitation.
\section*{Acknowledgment}
This work was supported by JST Moonshot R\&D Program, Grant Number JPMJMS223B. Financial support to A. De Luca and L. Nunziante from PNRR MUR project PE0000013-FAIR is also acknowledged. The authors would like to thank Hamano Products Co., Ltd.\ and Prof. Fumitoshi Matsuno's research group in Moonshot R\&D Program for their invaluable support in the development of the hardware platform, and are grateful to Pascal Pama for the assistance in conducting the experiments.
\vspace{-5pt}

\bibliographystyle{IEEEtran}
\bibliography{references.bib}

\end{document}

%% file: macros.tex
\newcommand{\fig}[1]{Fig.~\ref{#1}}

\def\epsgaiji#1{\leavevmode\kern-0.025zw\raise-.37zh\hbox{%
  \epsfile{file=#1,width=1.05zw}}\kern-0.025zw}
\newcommand{\MARU}[1]{{\ooalign{\hfil#1\/\hfil\crcr\raise.167ex\hbox{\mathhexbox20D}}}}

\usepackage{array}


%% file: references.bib
@inproceedings{Boucher2024,
author={Camille Boucher and others},
memo={Camille Boucher and Gustavo H. Diaz and Shreya Santra and Kentaro Uno and Kazuya Yoshida},
title={Integration of Vision-Based Object Detection and Grasping for Articulated Manipulator in Lunar Conditions},
booktitle = {Proc.\ IEEE/SICE Int.\ Symp.\ Syst. Integrat. (SII)},
pages ={484--489},
year = {2024},
doi={10.1109/SII58957.2024.10417086},
}

@article{Chien2024_fracture,
  title={{YOLOv9} for Fracture Detection in Pediatric Wrist Trauma {X}-ray Images},
  author={Chun-Tse Chien and Rui-Yang Ju and Kuang-Yi Chou and Jen-Shiun Chiang},
  journal={arXiv:2403.11249},
  year={2024}
}

@INPROCEEDINGS{Creech2022_Artemis,
  author={Creech and others},
  memo={Creech, Steve and Guidi, John and Elburn, Darcy},
  booktitle={Proc.\ IEEE Aerospace Conf.}, 
  title={Artemis: An Overview of {NASA}'s Activities to Return Humans to the Moon}, 
  year={2022},
  volume={},
  number={},
  pages={1--7},
  doi={10.1109/AERO53065.2022.9843277}
}

@inproceedings{Diaz2024_AROB,
author={Gustavo H. Diaz and others},
memo={Gustavo H. Diaz and Tharit Sinsunthorn and Shreya Santra and Kentaro Uno and Kazuya Yoshida},
title={Toward Autonomous Assembly of Modular Robots and Structures using Real-time Object Detection and Imitation Learning for Lunar Missions},
booktitle = {Proc. 29th Int. Symp. Artif. Life Robot. (AROB)},
pages ={1389--1393},
year = {2024},
doi={10.1109/SII58957.2024.10417086},
}

@inproceedings{Eppinger1986,
author={Eppinger, S. and Seering, W.},
title={On dynamic models of robot force control}, 
booktitle={Proc.\ IEEE Int.\ Conf.\ Robot. Automat.}, 
year={1986},
pages={29--34},
doi={10.1109/ROBOT.1986.1087723},
}

@inproceedings{Findeisen2002,
author = {Findeisen, Rolf and Allgöwer, Frank},
title = {An Introduction to Nonlinear Model Predictive Control},
booktitle={Proc.\ 21st Benelux Meeting on Systems and Control},
pages = {119--141},
year = {2002},
}

@article{Findeisen2003,
author={Rolf Findeisen and others},
memo={Rolf Findeisen and Lars Imsland and Frank Allgower and Bjarne A. Foss},
title={State and Output Feedback Nonlinear Model Predictive Control: An Overview},
journal={European Journal of Control},
volume={9},
number={2},
pages={190--206},
year={2003},
doi = {https://doi.org/10.3166/ejc.9.190-206}
}

@book{Isidori1995,
author={Isidori, Alberto},
title={Nonlinear Control Systems}, 
year={1995},
publisher={Springer},
address={London},
journal={Communications and Control Engineering},
doi={10.1007/978-1-84628-615-5},
}

@article{jin2022dope++,
  title={DOPE++: 6D pose estimation algorithm for weakly textured objects based on deep neural networks},
  author={Jin, Mei and Li, Jiaqing and Zhang, Liguo},
  journal={PloS one},
  volume={17},
  number={6},
  pages={e0269175},
  year={2022},
  publisher={Public Library of Science San Francisco, CA USA}
}

@misc{Jocher2023_Ultralytics_YOLO,
author={Jocher, Glenn and others},
memo={Jocher, Glenn and Chaurasia, Ayush and Qiu, Jing},
title={Ultralytics {YOLO}},
license={AGPL-3.0},
version={8.0.0},
month={Jan},
year={2023},
url={https://github.com/ultralytics/ultralytics},
}

@article{Khatua2024,
author={Khatua, Aniruddha and others},
memo={Khatua, Aniruddha and Bhattacharya, Apratim and Goswami, Arkopal K. and Aithal, Bharath H.},
title={Developing approaches in building classification and extraction with synergy of {YOLOV8} and {SAM} models},
journal={Spatial Information Research},
year={2024},
issn={2366--3294},
doi={10.1007/s41324-024-00574-0},
}

@book{Kouvaritakis2016,
author={Kouvaritakis, Basil and Cannon, Mark},
title={Model Predictive Control: Classical, Robust and Stochastic},
year={2016},
publisher={Springer},
address={Cham},
journal={Advanced Textbooks in Control and Signal Processing},
isbn = {978-3-319-24851-6},
doi={10.1007/978-3-319-24853-0},
}

@article{Krizhevsky2012_CommACM2017,
author={Krizhevsky, Alex and others},
memo={Krizhevsky, Alex and Sutskever, Ilya and Hinton, Geoffrey E},
title={{ImageNet} Classification with Deep Convolutional Neural Networks},
journal={Communication of the ACM},
volume={60},
number={6},
pages={84--90},
year={2017}
}

@inproceedings{Lin2017_CVPR,
author={Tsung-Yi Lin and others},
memo={Tsung-Yi Lin and Piotr Dollár and Ross Girshick and Kaiming He and Bharath Hariharan and Serge Belongie},
title={Feature Pyramid Networks for Object Detection},
booktitle={Proc. IEEE Conf. Computer Vision and Pattern Recognition (CVPR)},
pages={936--947},
year={2017},
}

@misc{Nikkei2024,
  title         = {{U.S.} and {J}apan spearhead new era of private moon landings},
  year          = {2024},
  memo          = {Accessed: 2024-03-21},
  howpublished  = {\url{https://asia.nikkei.com/Business/Aerospace-Defense-Industries/U.S.-and-Japan-spearhead-new-era-of-private-moon-landings}}
}

@article{Qin1997,
author = {Qin, Joe and Badgwell, Thomas},
year = {1997},
pages = {232-256},
title = {An Overview Of Industrial Model Predictive Control Technology},
volume = {93(1)},
journal = {AIChE Symp.\ Ser.},
}

@inproceedings{Redmon2016_YOLO,
author={Joseph Redmon and others},
memo={Joseph Redmon and Santosh Divvala and Ross Girshick and Ali Farhadi},
title={You Only Look Once: Unified, Real-Time Object Detection}, 
booktitle={Proc. IEEE Conf. Computer Vision and Pattern Recognition (CVPR)},
pages={779--788},
year={2016},
}

@article{rijal2023comparing,
  title={Comparing Depth Estimation of Azure Kinect and Realsense D435i Cameras},
  author={Rijal, Sanjay and Pokhrel, Suruchi and Om, Madhav and Ojha, Vaghawan Prasad},
  journal={Available at SSRN 4597442},
  year={2023}
}

@book{Siciliano2010,
author={Siciliano, Bruno and Sciavicco, Lorenzo and Villani, Luigi and Oriolo, Giuseppe},
memo={Siciliano, Bruno and others},
title={Robotics: Modelling, Planning and Control},
year={2010},
publisher={Springer},
address={London},
}

@article{Song2019,
author={Song, Peng and others}, 
memo={Song, Peng and Yu, Yueqing and Zhang, Xuping}, 
year={2019}, 
pages={801--836},
title={A Tutorial Survey and Comparison of Impedance Control on Robotic Manipulation},
volume={37},
DOI={10.1017/S0263574718001339},
number={5},
journal={Robotica},
}

@article{Wang2024_yolov9,
  title={{YOLOv9}: Learning What You Want to Learn Using Programmable Gradient Information},
  author={Wang, Chien-Yao and Yeh, I-Hau and Liao, Hong-Yuan Mark},
  journal={arXiv:2402.13616},
  year={2024}
}

@inproceedings{Xia2018,
author = {Xia, Gui-Song and others},
memo = {Xia, Gui-Song and Bai, Xiang and Ding, Jian and Zhu, Zhen and Belongie, Serge and Luo, Jiebo and Datcu, Mihai and Pelillo, Marcello and Zhang, Liangpei},
title = {{DOTA}: A Large-Scale Dataset for Object Detection in Aerial Images},
booktitle = {Proc. IEEE Conf. Computer Vision and Pattern Recognition (CVPR)},
pages ={3974--3983},
year = {2018},
}

@article{Xiao2020,
author={Xiao, Youzi and others},
memo={Xiao, Youzi and Tian, Zhiqiang and Yu, Jiachen and Zhang, Yinshu and Liu, Shuai and Du, Shaoyi and Lan, Xuguang},
title={A review of object detection based on deep learning},
journal={Multimedia Tools and Applications},
year={2020},
month={Sep},
day={01},
volume={79},
number={33},
pages={23729--23791},
issn={1573-7721},
doi={10.1007/s11042-020-08976-6},
}

@article{Xiao2024,
author={Xiao, Bingjie and others},
memo={Xiao, Bingjie and Nguyen, Minh and Yan, Wei Qi},
title={Fruit ripeness identification using {YOLOv8} model},
journal={Multimedia Tools and Applications},
year={2024},
volume={83},
number={9},
pages={28039--56},
issn={1573-7721},
doi={10.1007/s11042-023-16570-9},
}
